\documentclass[preprint,12pt]{elsarticle}

\usepackage{amssymb}
\usepackage{makecell}
\usepackage{rotating}
\usepackage[pagewise]{lineno}
\journal{Computers in Biology and Medicine}

\begin{document}

\begin{frontmatter}

%% Title, authors and addresses

%% use the tnoteref command within \title for footnotes;
%% use the tnotetext command for theassociated footnote;
%% use the fnref command within \author or \address for footnotes;
%% use the fntext command for theassociated footnote;
%% use the corref command within \author for corresponding author footnotes;
%% use the cortext command for theassociated footnote;
%% use the ead command for the email address,
%% and the form \ead[url] for the home page:
%% \title{Title\tnoteref{label1}}
%% \tnotetext[label1]{}
%% \author{Name\corref{cor1}\fnref{label2}}
%% \ead{email address}
%% \ead[url]{home page}
%% \fntext[label2]{}
%% \cortext[cor1]{}
%% \affiliation{organization={},
%%             addressline={},
%%             city={},
%%             postcode={},
%%             state={},
%%             country={}}
%% \fntext[label3]{}

\title{Deep Convolutional Neural Networks for Onychomycosis Detection}

%% use optional labels to link authors explicitly to addresses:
%% \author[label1,label2]{}
%% \affiliation[label1]{organization={},
%%             addressline={},
%%             city={},
%%             postcode={},
%%             state={},
%%             country={}}
%%
%% \affiliation[label2]{organization={},
%%             addressline={},
%%             city={},
%%             postcode={},
%%             state={},
%%             country={}}

\author[inst1]{Abdurrahim Yilmaz}

\affiliation[inst1]{organization={Department of Mechatronics Engineering, Yildiz Technical University},%Department and Organization
            %addressline={Address One}, 
            city={Besiktas},
            postcode={34349}, 
            state={Istanbul},
            country={Turkey}}

\author[inst2]{Fatih Göktay}
\author[inst1]{Rahmetullah Varol}
\author[inst3]{Gulsum Gencoglan}
\author[inst1]{Huseyin Uvet}

\affiliation[inst2]{organization={Department of Dermatology and Venereology, University of Health Sciences Turkey, Hamidiye Medical Faculty, Haydarpasa Numune Training and Research Hospital},%Department and Organization
            %addressline={Address Two}, 
            city={Uskudar},
            postcode={34668 }, 
            state={Istanbul},
            country={Turkey}}

\affiliation[inst3]{organization={Department of Dermatology, Liv Hospital Vadistanbul, Istinye University},%Department and Organization
            %addressline={Address Two}, 
            city={Sariyer},
            postcode={34396 }, 
            state={Istanbul},
            country={Turkey}}

\begin{abstract}
The diagnosis of superficial fungal infections in dermatology is still mostly based on manual direct microscopic examination with Potassium Hydroxide (KOH) solution. However, this method can be time consuming and its diagnostic accuracy rates vary widely depending on the clinician's experience. With the increase of neural network applications in the field of clinical microscopy, it is now possible to automate such manual processes increasing both efficiency and accuracy. This study presents a deep neural network structure that enables the rapid solutions for these problems and can perform automatic fungi detection in grayscale images without dyes. 160 microscopic field photographs containing the fungal element, obtained from patients with onychomycosis, and 297 microscopic field photographs containing dissolved keratin obtained from normal nails were collected. Smaller patches containing 4234 fungi and 4981 keratin were extracted from these images. In order to detect fungus and keratin, VGG16 and InceptionV3 models were developed. The VGG16 model had 95.98\% accuracy, and the area under the curve (AUC) value of 0.9930, while the InceptionV3 model had 95.90\% accuracy and the AUC value of 0.9917. However, average accuracy and AUC value of clinicians is 72.8\% and 0.87, respectively. This deep learning model allows the development of an automated system that can detect fungi within microscopic images.
\end{abstract}

%%Graphical abstract
%\begin{graphicalabstract}
%\begin{figure*}[ht]
%    \centering
%    \includegraphics[width=0.8\textwidth]{figures/ann.png}
%    \label{fig:ann_model}
%\end{figure*}%
%\end{graphicalabstract}
%
%%%Research highlights
%\begin{highlights}
%\item Onychomycosis classification with deep neural networks.
%\item The microscopic images of onychomycosis is used to detect fungis.
%\end{highlights}


\end{frontmatter}

%% \linenumbers

%% main text
\section{Introduction}
\label{sec:sample1}
Incorporation of deep learning methods with microscopy allows many applications such as microorganism classification, disease detection, segmentation, etc. \cite{xing2017deep}. Fungal diseases can be detected using microscopic images. Onychomycosis is also a fungal infection that occurs in 10\% of the total population and covers 50\% of nail diseases. There are many organisms that cause onychomycosis such as dermatophytes, Candida, and non-dermatophyte organisms. Dermatophytes (especially Trichophyton) are the most common with 90\% rate in the distribution \cite{westerberg2013onychomycosis}. Despite the technical advantages of traditionally used tests for the detection of the causative agents, none can be considered as a standard test alone from the viewpoint of their diagnostic utility. As a result, many criteria are frequently utilized in diagnostic validity studies since their use together can improve sensitivity and specificity \cite{stewart2012update}. However, there is currently no consensus on the most appropriate combination of tests because conflicting results have been reported for performance and validity. In this regard, studies have shown a high variability of results in the application of individual tests or their combinations. Sensitivity values of 11 clinical studies, regarding the manuel microscopic KOH examination methodology, have been reported in the range of 44\%-100\% with 61\% average value \cite{velasquez2017meta}. Another methodology that have been used in the clinic is the culture method where nail samples are cultured within an appropriate medium. However, the culturing period takes a very long time and there is a risk of laboratory contamination. The third methodology that is used during the clinic practice is histopathological examination with periodic acid schiff (PAS) staining. However, this methodology is more expensive and inconvenient for all patients. Compared to other methods, KOH examination is more practical and cheaper. However, it suffers from unreliable diagnostic performance due to dependence on the clinician experience. This problem resulted in wide range sensitivity rates. In this study, we want to improve practitioner related error rate and increase the sensitivity of the KOH examination method.

\begin{figure*}
    \centering
    \includegraphics[width=1\textwidth]{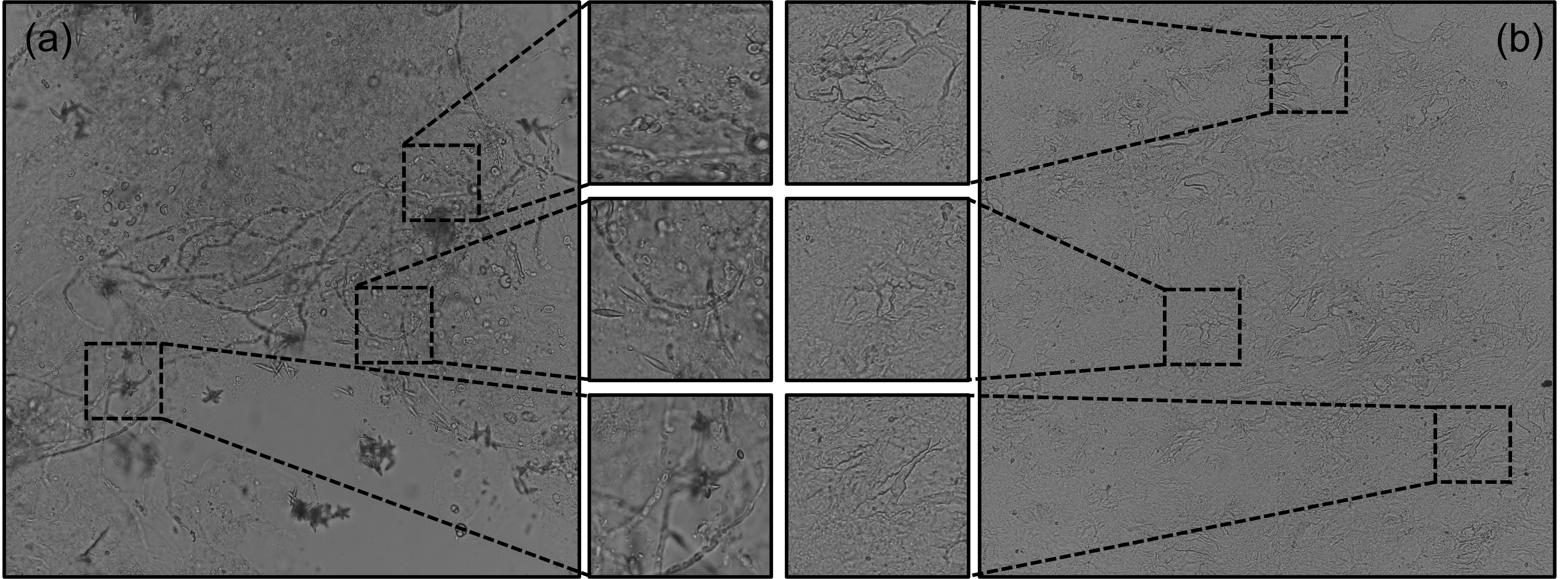}
    \caption{(a) Shows a microscopic image along with extracted patches that contain fungi. (b) Shows another microscopic image along with extracted patches that contain keratin. Microscopic images were converted to grayscale. All of the extracted patches were marked as either fungi or keratin. These patches were then used to train the deep neural network model.}
    \label{fig:example}
\end{figure*}%

Onychomycosis is a challenging disease to diagnose because sometimes it cannot be distinguished from other nail diseases like onychodystrophy. Patients must use the drugs for two months for fingernail onychomycosis and three months for toenail onychomycosis. Dermatologists want to be sure for exact diagnosis before starting anti-mycotic treatment because of their toxicity. Doctors may apply incorrect treatment if the decision is made by clinical features or wrong feedback from the mycological examination. So, clinicians depend on feedback from mycological examination for the management of the disease. In order to prevent incorrect diagnosis and reduce the workload of technicians and doctors, a reliable automated detection methodology must be developed. In this way, the false prediction rate is reduced because of workload.

In the diagnostic method of direct microscopic examination with KOH, diseased nail plate is dissolved inside a KOH solution. Afterwards, the sample is examined under a light microscope on a glass slide. Fungus are located in ceratin lamellas of nail that causes artifacts such as chaotic shapes in KOH solution, non-dissolved ceratin, crystals of KOH. For these reasons, to distinguish fungus is a challenging task. Dyes like Calcofluor White are used to improve time efficiency of examination and visibility of fungus patterns. This method needs a fluorescent microscopy and additional cost. Therefore, we aim to develop a fungus classification system to overcome challenge of direct KOH examination without using dyes.

By its nature, the dermatology has a very large database of images that can be used and interpreted with artificial intelligence. For this reason, it is a unique field of science that stands out among the artificial intelligence applications that can be used in the future. In dermatology, artificial intelligence can be used to in diagnosing and treatment of melanoma and nonmelanoma skin cancers, psoriasis and atopic dermatitis, etc. There are many studies investigating its use in diagnosis and treatment monitoring \cite{hogarty2020artificial} To the best of our knowledge, only one study has investigated the use of artificial intelligence for the diagnosis of onychomycosis so far \cite{han2018deep}. This study was carried out on the clinical photographs of the patients. Achieving standardization in clinical photography is very difficult. Diagnostic reality may be affected by the use of photographs with insufficient light and noise in artificial intelligence models.  

Main contributions of this study are; (1) the artificial intelligence-based method that can serve as a clinical decision support system in order to increase onychomycosis detection performance; (2) a method for onychomycosis detection in grayscale microscopic images that doesn't rely on dyes; (3) a real-time automated onychomycosis detection ability. Our research is the first to implement an Artificial Intelligence (AI) for detecting onychomycosis via microscope images. A Convolutional Neural Network (CNN) based deep learning algorithm is developed to locate fungus in microscopic images. In addition to the goal of scanning to find the fungus causing nail infections, we also intend to establish a coherent detection standard for clinical purpose. Two separate models were trained using patches obtained from microscopic images. The VGG16 and InceptionV3 structures were chosen as baseline architectures and the results were compared with the developed custom model as well as with clinician performance. 

\section{Methods/Theory}
Machine learning and in particular supervised learning has been used for many years in subjects such as disease detection, image classification, and segmentation \cite{erickson2017machine}. In the supervised learning approach, the deep learning model is trained gradually by comparing the outputs of the model with expected results. At each step, model parameters are updated such that the total error will be minimized. It has been shown in many studies that the supervised learning approach is the most robust training method which model complex relations between the input and the output space using ground truth data that is marked by an expert.

Algorithms in supervised learning are generally grouped under two classes as classification and regression. The classification method deals with problems in which a number of samples need to be distinguished according to their respective classes. Within this scope, deep learning stands out compared to traditional algorithms \cite{wang2021comparative}. The ability of deep neural networks to abstract very complex structures in intermediate layers and to capture patterns that humans cannot capture under normal conditions through stronger models stands out as an important advantage in terms of this problem \cite{geirhos2017comparing}. Thus, classification can be made according to the type, size, and shape of the input images. In this study, we use the supervised learning methodology in order to train two deep neural networks using patches of fungus and keratin that were extracted from microscopic images taken from a large number of slides.

\subsection{OnI (Onychomycosis Images) and NnI (Normal Nail Images) Dataset}
Data collection and preprocessing directly affect the success of the deep learning model \cite{jeong2018image}. To achieve optimal performance, data should be collected with minimal exposure to artificial disturbances such as environmental factors. In addition, the collected data must be systematically pre-processed to remove noise and artifacts from the images. While creating the dataset, clean, artifact-free images were collected in grayscale as shown in Fig. \ref{fig:example}. Ethical approval was taken from Institutional review board of Haydarpaşa Numune Training and Research Hospital (reference number: HNEAH-KAEK 2021/209).

Five cases who applied to the dermatology outpatient clinic with nail related complaints, were thought to be distal lateral onychomycosis in clinical and dermoscopic examination, and had positive KOH examination results of nail samples were included in the study. For KOH examination, after the affected nail plate was wiped with 70\% ethyl alcohol, a piece of the free end of the nail plate was cut with the help of a sterile nail nipper, moving as far as possible proximally. The sample taken was dissolved in 20\% KOH solution for 24 hours. The slides prepared from this molten sample were examined directly microscopically under a binocular Olympus CX23 microscope with 40X magnification. During the examination, the condenser of the microscope will be positioned closest to the slide and the diaphragm has been adjusted to the appropriate setting for 40X magnification. The results of the cases in which septal hyphae were detected in the direct microscopic examinations evaluated by a dermatologist expert in nail diseases (FG) with 20+ years of experience were accepted as positive. In the preparations of each positive case, multiple photographs were taken from different areas with hyphae structures entering the microscopic field. The photos were taken with a 24.2 megapixel camera with an 18-55 mm original lens attached to the Canon EOS M3 (Canon, Tokyo, Japan) brand, which is fixed with a tripod to the microscope ocular. The photos were taken in manual mode, with standard ISO 100, f/3.5 and 18mm focal length at 6000X4000 pixels. If necessary according to the light level, the adjustment was made by changing the shutter levels. As a control group, 5 cases who applied to the dermatology outpatient clinic for reasons other than nails and accepted to give a sample from the free edge of their healthy nails for KOH examination were included. Sampling, KOH examination and microscopic photographing of healthy nails of the control group were performed with the same methods applied for nails with onychomycosis. Microscopic photographs were taken from multiple different microscopic fields containing dissolved keratin. 160 images containing fungus and 297 images containing only keratin were collected. 9215 patches of 500$\times$500 pixels were collected from these images. In addition, patches containing more than 50\% of the image in the regions of the microscope edge in microscopic images are included in the dataset. Dataset statistics are shown in Table I.

\begin{table}[t!]
\renewcommand{\arraystretch}{1.25}
\centering
\label{tab:dataset}
\caption{Dataset Specifications}
\begin{tabular}{ccc} 
\hline\hline
\textbf{}  & \textbf{No. of Images} & \textbf{No. of Patches}             \\ 
\hline \\ [-10pt]
Fungus Images       &  160 & 4234 \\ [5pt]
Keratin Images  & 297 & 4981 \\ [5pt]
\hline \\ [-10pt]
Total & 457 & 9215 \\ [5pt]
\hline\hline
\end{tabular}
\end{table}

\begin{figure*}
    \centering
    \includegraphics[width=0.8\textwidth]{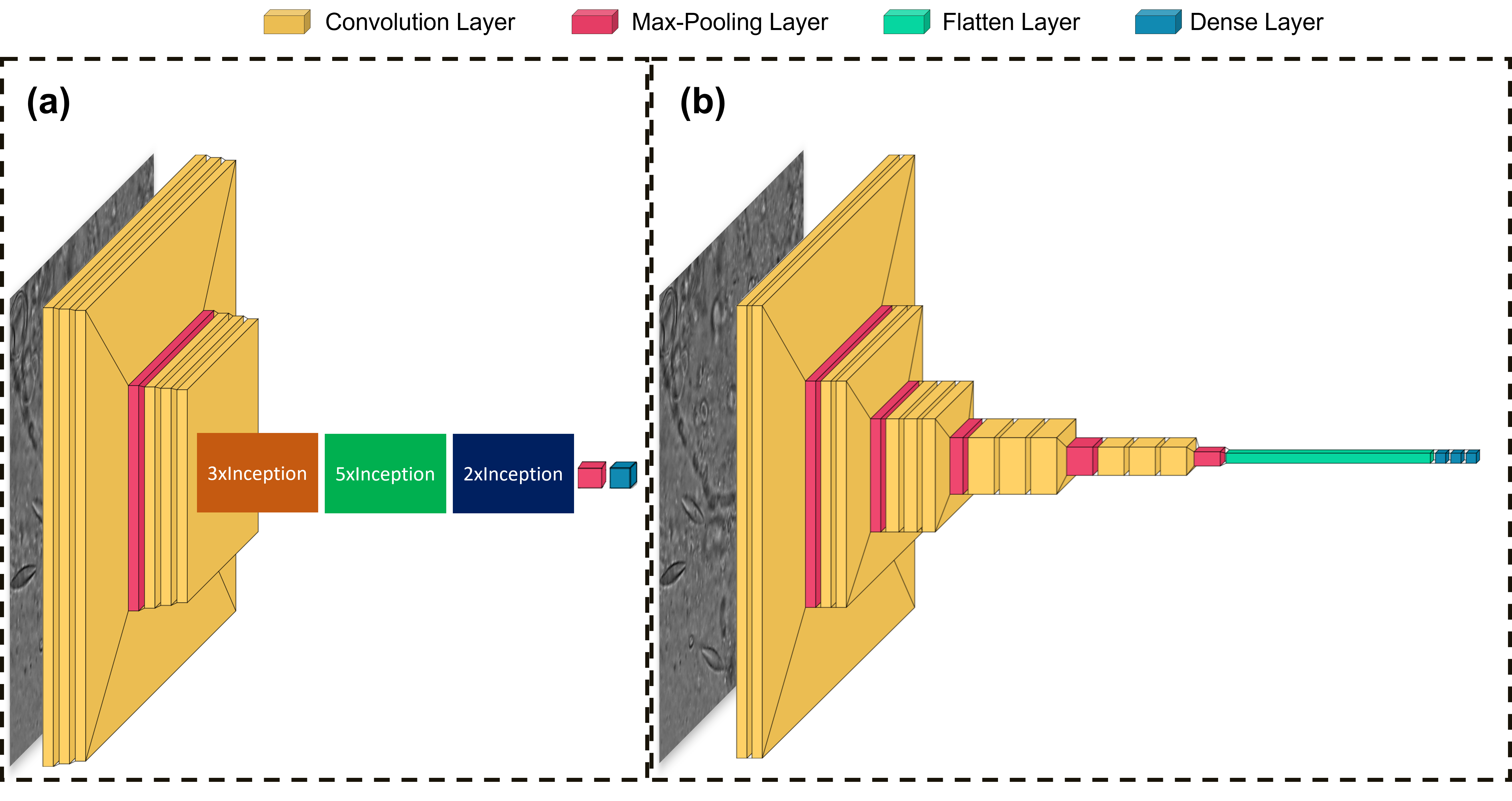}
    \caption{(a) Shows InceptionV3 model for fungus classification. The model includes two convolution blocks that consist of three convolution layer and one max pooling layer. After convolution layers, there is 3xInception, 5xInception, and 2xInception blocks. At the end of neural network, a fully connected structure which consists of one dense layer. (b) Shows layers of VGG16 deep learning architecture. The model includes two convolution blocks that consist of two convolution layers and one max pooling layer, and three convolution blocks that consist of three convolution layers and one max pooling layer as connected serially. After convolution layers, there is a fully connected structure which consists of two dense layers. The numbers in the boxes indicate the size of the image after the layer. Figures were created with visualkeras library \cite{gavrikov}.}
    \label{fig:ann_model}
\end{figure*}%

\subsection{Fungus Classification System}
The artificial neural network layers to be established with the help of this collected data set were as follows.
Firstly, the input layer is the layer where microscopic images were fed into the neural network. Convolution layers are where automated feature extraction from microscopic images were made. The weight and bias parameters of the nodes at the convolution layers are updated using the backpropagation algorithm such that the features that enable the model to distinguish the samples are learned. In contrast, classical image processing and machine learning algorithms are able to learn only handcrafted features which may be insufficient in capturing subtle details. In addition, by visualizing these convolution layers, it is possible to understand the decision making mechanism of the neural network. The pooling layer is the layer where the previous layer is downsized and the parameters and computational load are reduced. The layers up to this point are the parts of the artificial neural network where feature extraction and preprocessing are done. In order to establish a relationship between features, it is necessary to learn which feature is associated with which category. In order to establish this relationship, the features are brought into a single line and made suitable for the input of the fully connected neural network structure in the flattening layer. After the automated feature extraction, fully connected neural network layers are where the relationship between features is established. Finally, the output layer is the layer in which the probabilistic value of each category is obtained \cite{lecun1995convolutional}.

\subsubsection{InceptionV3 Model}
The first neural network structure used is InceptionV3. Although this network has much less parameters than VGG16 architecture, it can achieve high accuracy and sensitivity, approximately equal to VGG16 architecture. As in the VGG16 architecture, images are scaled to 224x2224x3. The image passes through each block layer until it reaches the max-pooling layer on the next block layer. The image passes via the convolution blocks and into the flatten layer, which is fully linked. One dense layer with 64 nodes and one dense layer with 32 nodes as a fully connected layer were produced in this work to establish a relationship between fungus and keratin patch traits. The softmax activation function is used in the InceptionV3 structure's output layer. The created InceptionV3 structure is shown in Fig. II-(a).  

\subsubsection{VGG16 Model}
VGG16 architecture was used for onychomycosis classification \cite{simonyan2014very}. Unlike its predecessor neural network structures, it uses convolution layers in 2 or 3 ways. Therefore, it contains many parameters. This causes the training of the network to take too long and makes fine-tuning difficult. In order to use the VGG16 structure, images must be scaled to 224x224x3 on input. Therefore, grayscale images are given to the network as 3 channels instead of a single channel. After the input layer, there are 2 double convolution layers and then 3 triple convolution layers. The image passing through each block layer reaches the next block layer with the max pooling layer. The image passing through the convolution blocks reaches the fully connected layer with the flatten layer. In this study, 1 dense layer with 64 nodes, and 1 dense layer with 32 nodes as a fully connected layer, which is used to establish a relationship between features of fungus and keratin patches, were created. The output layer in the VGG16 structure uses the softmax activation function. The created VGG16 structure is shown in Fig. II-(b).

\subsubsection{Computational Speed}
Models were trained using computer with 1080Ti GPU and 32GB Ram. For the VGG16 structure, training was conducted for 200 epochs and took 160 minutes. For the InceptionV3 model, training was conducted for 200 epochs and took 170 minutes. The average test time for a patch is 3.6207 milliseconds and 3.5842 milliseconds for VGG16 and InceptionV3, respectively.

\section{Results}
\begin{figure}[t]
    \centering
    \includegraphics[width=0.8\columnwidth]{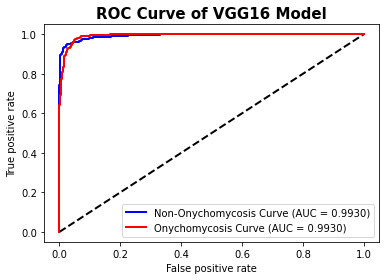}
    \caption{Shows ROC Curve of VGG16 model with AUC value.}
    \label{fig:vgg_roc}
\end{figure}

\begin{figure}[t]
    \centering
    \includegraphics[width=0.8\columnwidth]{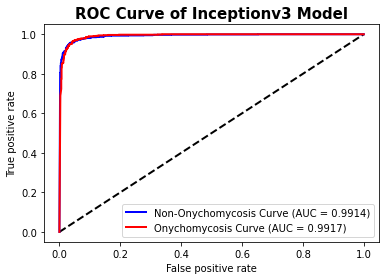}
    \caption{Shows ROC curve of InceptionV3 model with AUC value.}
    \label{fig:inc_roc}
\end{figure}

For supervised deep learning models, a dataset called test data is used as the control group. Test data is randomly selected from a portion of the previously acquired dataset. Thus, model metrics can be calculated without the need to collect data again. There are many classification metrics such as accuracy, which measures the overall performance of the model, and $F_1$ Score, which measures positive and negative prediction performances. However, since such classical metrics are calculated with regards to a threshold value, they may not be sufficient to express the performance of the model. Therefore, the use of metrics such as receiver operating characteristics (ROC) curve and Area Under the Curve (AUC) which show the model performance for different threshold values, are effective in better understanding the model performance. The ROC curves of the models are shown in Fig. \ref{fig:vgg_roc} and \ref{fig:inc_roc} for VGG16 and InceptionV3 models respectively. According to the results from the meta-analysis of 11 clinical studies, the average success of the experts in fungi identification is 72.8\% and the AUC value is 0.87 \cite{velasquez2017meta}.

\begin{sidewaystable*}
\renewcommand{\arraystretch}{1.7}
\centering
\caption{Results for Models and Average Values of Clinician for KOH Examination}
\begin{tabular}{ccccc} 
\hline\hline
\textbf{Metric} &\textbf{Describtion} & \textbf{InceptionV3}  & \textbf{VGG16}   & \textbf{Average Values for Clinician \cite{velasquez2017meta}}          \\ 
\hline \\ [-10pt]
Accuracy &$\frac{T P+T N}{T P+F P+T N+F N}$   &  95.90\%  &  95.98\% & 72.8\% \\ [5pt]
Sensitivity
&$\frac{T P}{T P+F N}$ & 95.50\%  & 95.98\% &  61\% \\ [5pt]
Precision &$\frac{T P}{T P+F P}$ & 95.58\%  & 96.03\% & 96.3\% \\[5pt]
Specificity &$\frac{T N}{T N+F P}$ & 97.50\%  & 97.53\%  & 95\% \\ [5pt]
$F_1$ Score &$2\times\frac{Precision \times Sensitivity}{Precision+Sensitivity}$ & 95.50 & 95.99 & 74.69 \\ [5pt]
AUC & Area Under the ROC Curve & 0.9917 & 0.9930 & 0.87 \\ [5pt]
Total Parameter & Parameter Number of Models & 138,357,544  & 23,851,784 & - \\ [5pt]
Speed (s) & Inference Time of Test Data & 3.5842  & 3.6207 & - \\ [5pt]
\hline\hline
\end{tabular} \label{tab:results}
\end{sidewaystable*}

ROC curve and AUC metrics were used together with the classical metrics that accepted the 0.5 threshold value to measure the performance of the model. Also, the parameter count and testing time for a single patch were compared. 1843 patches, which make up 20\% of the total patches, were used for test data. During model training, 15\% of the patches were used for validation. As per the result of the developed models shown in Table \ref{tab:results}, VGG16 model also found a better estimation with 95.98\% success than InceptionV3 model with 95.90\% success. VGG16, which resulted in a precision value of 95.98\%, was more successful than the InceptionV3 model which resulted in a precision value of 95.50\% in positive fungus prediction.

\section{Discussion}
Recently there have been a number of studies where a deep learning model was shown to classify better than human experts \cite{liu2019comparison}. The utilization of such models in the clinic will decrease the workload on experts and erroneous diagnoses. Furthermore, since expert performance is a parameter that increases with experience, there is a big difference in performance between the performance of less experienced experts and those who have worked in the field for many years. In addition to experience, the average number of sample views also highly contributes to success \cite{bunyaratavej2016experiences}. The use of decision support systems can help clinicians with less experience to make diagnoses with high reliability and achieve a performance closer to that of an experienced clinician.

One of the biggest problems in the detection of hyphae in the KOH slides is that a large area of the sample needs to be scanned in order to find the fungus or ensure their absence. This time can be reduced by adding dyes into the KOH preparations. However, these dyes are not available in some cases or clinics. In this study, a model was developed that allows automatic fungal detection thanks to microscopic images collected without staining, containing only keratin and fungi. The collected images are not microscope-dependent as they were collected under different light conditions and background colors. In order to develop a robust model that can perform in the presence of various noises, the model was trained by converting the images to grayscale which eliminates confusion due to color and light differences. Furthermore, for the model to be used in real-time in clinical settings it needs to be able to classify rapidly. The VGG16 and InceptionV3 models are appropriate to be used in a real-time fashion for the large number of patches that need to be extracted from a single microscopic image.

Biopsy, histopathology, mycological culture, KOH examination, polymerase chain reaction (PCR), etc. are among the tests that can be used to diagnose onychomycosis \cite{gupta2013diagnosing}. The most often utilized procedures are direct KOH examination, culture, and nail plate biopsy since some of these tests are highly expensive, need specialized equipment and supplies, and take a long time and are not consistently employed. Although there are many diagnostic studies with KOH tests in both onychomycosis and fungal infections of the skin, this test is an old and well-known methodology. The sensitivity of the method remains at the level of 44-100\% with 61\% average value. This high variation in the success rate is related to the expertise level of the clinician. In comparison, the VGG16 model have 95.98\% sensitivity which indicates a better performance compared to the average clinician. In addition, clinicians have low false positive values whereas false negative values are high. Therefore, it is a common occurance that unnecessary treatments are prescribed for non-fungal patients which poses a great health risk. The deep learning model, on the other hand results in low false positive rates and low false negative rates. This will be a useful feature in preventing unnecessary onychomycosis treatments. 

\section{Conclusion}
This paper presents a convolutional neural network-based onychomycosis classification technique using grayscale microscopic images. To the best of our knowledge, it is the first study in which a deep learning model is used for the purpose of onychomycosis detection. VGG16 and Inception V3, which are the successful deep neural network architectures, were trained, and the classification speed of these models were found to be sufficient for real-time applications. The performance of the developed model was compared to that of clinicians. Thus, this study revealed a pioneering method for reliably detecting onychomycosis. Additionally, a research-specific collection of images depicting onychomycosis and keratin has been published. Subsequent studies may yield additional and better models when the multiple instance learning and image segmentation methods are applied to the dataset.

%% If you have bibdatabase file and want bibtex to generate the
%% bibitems, please use
%%
 \bibliographystyle{elsarticle-num} 
 \bibliography{cas-refs}

%% else use the following coding to input the bibitems directly in the
%% TeX file.

% \begin{thebibliography}{00}

% %% \bibitem{label}
% %% Text of bibliographic item

% \bibitem{}

% \end{thebibliography}
\end{document}